\documentclass{article}

\usepackage[preprint]{nips_2018}

\usepackage[utf8]{inputenc} 
\usepackage[T1]{fontenc}    
\usepackage{hyperref}       
\usepackage{url}            
\usepackage{booktabs}       
\usepackage{amsfonts}       
\usepackage{nicefrac}       
\usepackage{microtype}      
\usepackage{graphicx}
\usepackage{subfigure}
\newcommand{\dnote}[1]{{}}
\newcommand{\snote}[1]{{}}
\title{On the Blindspots of Convolutional Networks}

\author{
  Elad Hoffer \\
    Technion - Israel Institute of Technology \\
    Haifa, Israel \\
  \texttt{elad.hoffer@gmail.com} \\
   \And
   Shai Fine \\
   The Interdisciplinary Center \\
   Herzliya, Israel \\
   \texttt{shai.fine@idc.ac.il} \\
   \And
   Daniel Soudry \\
   Technion - Israel Institute of Technology \\
   Haifa, Israel \\
   \texttt{daniel.soudry@gmail.com} \\
}

\begin{document}

\maketitle

\begin{abstract}
  Deep convolutional network has been the state-of-the-art approach for a wide variety of tasks over the last few years. Its successes have, in many cases, turned it into the default model in quite a few domains. In this work, we will demonstrate that convolutional networks have limitations that may, in some cases, hinder it from learning properties of the data, which are easily recognizable by traditional, less demanding, models.
  To this end, we present a series of competitive analysis studies on image recognition and text analysis tasks, for which convolutional networks are known to provide state-of-the-art results. In our studies, we inject a truth-revealing signal, indiscernible for the network, thus hitting time and again the network's blind spots. The signal does not impair the network's existing performances, but it does provide an opportunity for a significant performance boost by models that can capture it.
  The various forms of the carefully designed signals shed a light on the strengths and weaknesses of convolutional network, which may provide insights for both theoreticians that study the power of deep architectures, and for practitioners that consider applying convolutional networks to the task at hand.
\end{abstract}

\section{Introduction}
\label{introduction}
Over the last few years, deep convolutional networks (ConvNets) demonstrated exceptional performances, many a time near human-level, in a wide range of tasks.
These successes have made ConvNet the model of choice in quite a few domains.
What makes ConvNets so successful is less obvious, and a considerable effort is directed to study this question.
This work is a modest contribution to this effort.

We will show that ConvNets have limitations that may,
in some cases, impair their ability to learn critical properties of the data,
properties which are easily recognizable by traditional, less demanding, models.

Two recent studies have addressed the question what may cause deep neural networks (DNN) to fail measurably comparing to human vision:
\cite{szegedy2013intriguing} showed that changing an image in a way imperceptible to humans can cause a DNN to label the image as something else entirely (i.e., False Negative),
and in work by \cite{nguyen2014deep} it was shown that it is easy to produce images that are completely unrecognizable to humans,
but state-of-the-art DNNs believe that these are recognizable objects with 99.99 confidence (i.e. False Positive).

In this study, we take a different approach and rather than making the network completely fail,
we examine cases in which the network fails to capture significant information embedded in the data.
To do this, we inject highly relevant information in objects which are known to be recognizable by ConvNets.
This additional information does not derogate the network's existing performances. However, if captured, it may boost the performance significantly.
In all cases, the ConvNet models, which achieved best performances beforehand, were not able to improve performance by exploiting this additional information.
On the other hand, commonly used models, such as decision trees, random forest, Naive Bayes, logistic regression, and even shallow neural networks, which were all significantly inferior to ConvNets beforehand, were able to surpass the ConvNet's best performances by utilizing the embedded information.

The goal of this exercise is to gain a deeper understanding of the strengths and weaknesses of ConvNets. We will do it by comparing and matching the ConvNet performances with the competing models. Projecting the behavior of the various models over the carefully designed use cases while considering data characteristics, should teach us about modeling properties of ConvNets, and provide additional insights of their strengths and limitations.
We will argue that these limitations should be taken into account, and that data characteristics can greatly affect the applicability of ConvNets.


The rest of the paper is organized as follows: In section \ref{sec_blindspot} we review ConvNets, discuss their properties, and explain the design of truth revealing signals that hit ConvNets blindspots. Section \ref{sec_experiment} detail the comparative studies conducted in image classification (c.f. \ref{subsec_vision}) and text classification (c.f \ref{subsec_text}) tasks. Section \ref{sec_discussion} conclude with a discussion on the implications of the empirical results, lesson learned, and future research directions.

\section{Previous works}
\label{sec_previous_work}

In a previous study, \cite{szegedy2013intriguing} were able to show that it is easy to modify images in a way imperceptible to human vision, such that a trained network will fail
to classify them. It was also shown that certain types of noise, learned through optimization in the image space, are causing miss-classification 
on different networks and different datasets as well. This indicates that ConvNets are reliant on certain aspects of natural-image data,
that when modified, causes them to fail. 
\dnote{A few things: (1) this repeats a bit the information in the intro, which was only a few paragraphs ago. Maybe move section to end of paper before the discussion? (2) In the last sentence you use both "Convnets" and "convolutional networks". Better choose one name and stick to it throughout paper.}
\snote{fixed (2). As for (1) - I agree but I think it is minor and we can keep it as is}

In a follow-up work by \cite{goodfellow2014explaining} it was argued that these phenomena are related to quantization properties of natural images - The quantization used for pixel values creates ''gaps'', which when optimized against classification objective, causes the network to fail. Adversarial examples, in essence, exploit these ''gaps'' in the input space, leveraging unperceived (by humans) artifacts that do not exhibit characteristic traits of natural data. Generating adversarial examples and incorporating
them in the training procedure can improve the generalization on the entire test set, and reduce the error on the adversarial examples. However, additional examples can be created using the subsequent model.
Even when explicitly trained to avert adversarial examples, networks were shown to be susceptible to other kinds of noise \citep{carlini2017adversarial}. These adversaries were also expanded to include even single-pixel attacks \citep{su2017one}.

Complementary work showed 
a variety of geometrical shapes. These shapes are interpretable by humans; however, they share little resemblance to the classified category.
Nguyen et al.'s work highlights the limitations of deep ConvNets arising from their generalization capability --- it makes use of 
visual features that were learned to discriminate between training samples, to create pseudo images that fooled the trained network.

Another recently shown \citep{shalev2017failures} vulnerability of these models is their reliance on back-propagation \citep{rumelhart1986learning} and gradient descent methods to optimize their parameters. Shortcomings of these algorithms can hinder the model ability to learn otherwise simple tasks. In contrast, our work shows that besides limitations induced by training mechanism (such as the use of SGD), ConvNets also exhibit issues that stem from their structure and priors they impose.

\section{ConvNets blindspots}
\label{sec_blindspot}

\subsection{Convolutional networks}


Deep convolutional neural network is the leading approach of deep learning in computer vision tasks.
The main premise is that stationary properties of images allow the use of a reduced set of parameters that needs to be learned.
Convolutional layers consist of a set of trainable weight kernels, that produce their output value by cross-correlating with input data. This way, every spatial patch in the image is weighted using the same shared kernels. In each layer, multiple values are describing each spatial location. The different values for each location are known as the layer's \emph{feature maps}.


Since convolutional layers contain weights that are shared by multiple spatial regions, they naturally aggregate and average the gradients over large amounts of data. 
The same function can be applied as 1-dimensional convolution that can be applied to temporal data such as audio and text.

Another main layer of ConvNets is the pooling layer. Pooling layers are used to reduce the dimensionality of the input while gaining scale and shift invariance to small amounts.
This is done by simply pooling (or down-sampling) spatial regions of the input, taking the average (Average-Pooling), max (Max-Pooling) or other variants.

Most ConvNets are comprised of layers of Convolution \& Pooling followed by fully-connected layers (dot products) and a classifier.
Between layers with learned parameters, a non-linearity function is applied. Modern deep learning models usually employ 
\emph{ReLU} (rectified-linear-unit) $f(x)=\max(0,x)$ or some variant of it \citep{he2015delving}.
Convolutional networks are used for different vision-related tasks such as classification \citep{krizhevsky2012imagenet},
semantic segmentation \citep{long2015fully}, detection \citep{Sermanet} and control \citep{mnih2013playing}.

ConvNets were also shown to provide competitive results in language domain on tasks such as translation \citep{kalchbrenner2016neural} and classification \citep{zhang2015character}
by using convolutions on a character level. 

Convolutional networks are trained by variants of stochastic-gradient-descent (SGD) on batches, and usually, contain a number of parameters that
exceed the number of data samples by a large margin (over-specification). 
Another noteworthy aspect of ConvNets is that the input is processed over local regions (kernel sp sizes are usually on the scale of 3-11 pixels), while global
information is gathered by stacking multiple layers so that the final "spatial-region" of the model is at the scale of the whole image/sample.

Lately, residual connections, first introduced by the ResNet model by \cite{he2016deep} have been widely adopted across different deep architectures \citep{wu2016google, xie2017aggregated}. Residual connections add activations across layers (e.g $x^{\ell+1}=F(x^\ell) + x^\ell$ for a given layer $\ell$) and noted to improve convergence and final accuracy, as well as to enable the training of very deep networks. The merits of residual connections can be explained by their ability to allow gradients to "flow" throughout the network unhindered, which is significant for some of our findings and claims.

It is known that structure decisions and assumptions in the design of ConvNets are priors that form strong inductive bias concerning their input data \citep{cohen2016inductive, ulyanov2017deep}. We will try to examine and tackle some of these assumptions in comparison to other machine learning models.

\subsection{Injecting truth-revealing signal}

The use of convolutional networks is based on the assumption that the data fed into the network has
stationary properties, with strong locality either in space (vision) or time (text, speech, etc.).
While providing a strong prior, which is helpful for many natural data sources, this assumption may also lead to ``blindspots'' 
that should be reckoned with:
\begin{itemize}
 \item Inability to identify \emph{low-dimensional} signals in the input space, due to aggregation and ``smoothing'' by averaging of many data points
 \item Inability to identify \emph{global} signals within the data, due to the focus on local features 
\end{itemize}
The third type of ''blindspots'', which we term \emph{moving target}, is comprised by a combination of the former two
\begin{itemize}
 \item Inability to identify low-dimensional signals which vary in space and time at a global scale, due to the non-stationary nature of that signal
\end{itemize}

These ''blindspots'' are not necessarily harmful in data such as natural images, meaning they may not interfere with the learning and scoring conducted by ConvNets (e.g. unlike the signals described in Section \ref{sec_previous_work}). However, failing to capture the information embedded in the "blindspots" may cause convolutional networks to miss out information which is crucial for the task at hand.

To study these limitations of convolutional nets, we designed truth-revealing signals that reside well within the "blindspots", and embedded these signals in the data. 
The amount of information injected to the data varied from incomplete to complete, i.e., from lossy to a lossless encoding of the learning targets (the classification labels). Nevertheless, in all our case studies, the information conveyed by the embedded signals was significant, and models that could capture and exploit these signals were able to substantially improve their accuracy.

\section{Experiments}
\label{sec_experiment}
 \subsection{Vision - image classification}
 \label{subsec_vision}
 
\begin{figure}[h]
\begin{minipage}{.5\textwidth}
\begin{center}
\includegraphics[width=0.4\linewidth]{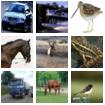}
\end{center}
   \caption{Original images}\label{fig_orig_img}

\end{minipage}
\begin{minipage}{.5\textwidth}

\begin{center}
\includegraphics[width=0.4\linewidth]{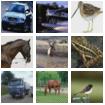}
\end{center}
   \caption{Noisy images}\label{fig_noisy_img}

\end{minipage}
\end{figure}

All the experiments were conducted on the CIFAR-10 dataset\citep{Krizhevsky2009cifar10}, which consists of 60000 32x32 color images in 10 classes, with 6000 images per class. There are 50000 training images and 10000 test images (c.f Fig \ref{fig_orig_img}).

\begin{figure}[h]
\begin{minipage}{.3\textwidth}
\begin{center}
\includegraphics[width=0.6\linewidth]{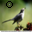}
\end{center}
\caption{One pixel encoding}\label{fig_one_pixel}
\end{minipage}
\begin{minipage}{.3\textwidth}
\begin{center}
\includegraphics[width=0.6\linewidth]{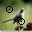}
\end{center}
\caption{Multiple locations}\label{fig_multiple_locations}
\end{minipage}
\begin{minipage}{.3\textwidth}
\begin{center}
\includegraphics[width=0.63\linewidth]{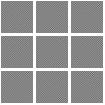}
\end{center}
\caption{Orthogonal noise}\label{fig_orth_noise}
\end{minipage}
\end{figure}

In our studies, we employed six different coding schemes of the ground-truth labels in the images:
\begin{itemize}
 \item One-pixel encoding: Changing one pixel (same location for all images) to hold the label (class number). An example of a bird's image with the encoded pixel (circled) is depicted in Figure \ref{fig_one_pixel}.
 \item Pattern pixels encoding: Encoding the label by changing the pixels at specific fixed locations to hold the class number (label). In our study, we used the four corners of the image for a binary encoding scheme, such that 10 labels are encoded with 4 pixels (1 in each corner), each of which represent a bit assignment, resp.
 \item Random pixel encoding: For each instance (image), encode the label (class number) in a randomly chosen pixel
 \item Multiple locations encoding: For each instance (image), encode the label with a pair of pixels, at distant locations, constrained to hold the exact same value. Thus, for every class, the label is encoded by the same location of the pair pixels that hold the same value, but the value itself changes from image to image. An example of the location encoding, where the pair pixels are circled, is depicted in Fig. \ref{fig_multiple_locations}.
 \item Noise encoding: The label is encoded by adding a random noise vector for each class. Those random vectors are dense and orthogonal to each other and with a large enough norm. Thus, the noisy vectors (cf. Fig. \ref{fig_orth_noise}) are in essence a lossless encoding of the labels, while the image itself (which is only mildly distorted, cf. Fig. \ref{fig_noisy_img} compared to the undistorted images in Fig. \ref{fig_orig_img}) holds only partial information\footnote{From a geometric standpoint, the encoding is a translation of the instances of each class along a different orthogonal axis. The translation preserves local geometric properties, thus making the scale and shift invariance properties less appealing, as the information is in the direction of translation}.
 \item Mean encoding: the label is encoded within the mean value of an image. We ensure the mean of each image is zero, and add to it the class number times a small coefficient of $1e-3$ so that computing it's new mean value can identify its label. 
\end{itemize}

\dnote{combine figure 5 with figure 1 (which is not very important by itself), or it would be hard to see the difference.}
\snote{Keeping figure 5 makes sense to me if we keep also figure 4. The value is to show that although we add the orthogonal noise, it doesn't affect the figures and the look the same also to the human eye. In that sense, the fact that figure 1 & 5 look very similar goes in favor of demonstrating this point. I emphasized it in the relevant bullet}

All of the above signals provide complete information, where the first and second are \emph{low dimensional} signals, the third and fourth are \emph{moving target} signals and the $5$th and $6$th are a \emph{global} signals. We note that the signals were encoded with scales considerably lower than original image ($1e-3$) to avoid being detected by simple weighting schemes.

In all the experiments we used a ResNet model \citep{he2016deep} which is widely used in visual classification problems.  We chose the original 56 layered version of the residual convolutional network with $~855K$ parameters.  A \emph{ReLU} non-linearity was used between consecutive layers as well as batch-normalization \citep{ioffe2015batch} according to the original work.
Training followed the regime of the original work by \cite{he2016deep} without data augmentation, as this will render most of our label-revealing signals completely undetectable.


The network achieved 84.9\% accuracy on the clean data, with no embedded signal, and with no data augmentation.
This result, although behind state-of-the-art results with data augmentation ($\sim 97\%$) \citep{gastaldi2017shake}, is a solid representative of a convolutional net on this task,
and suffices to demonstrate our points. 
Applying the network in all six truth-revealing signal did not yield any change in the network performance, for better or for worse. We do note that the use of residual connections significantly improved the network's ability to detect several additional attempts compared to networks with no residual connections. This may further explain some of the success of ResNet models, possibly allowing a higher signal-to-noise ratio in the gradients of consecutive layers.

For traditional modeling we used Python's scikit-learn package implementation  \citep{scikit-learn} in a straightforward manner, namely no special data pre-processing, no code enhancements, nor any tailor-made augmented functionality.
The models that were tested are - GaussianNB, Decision Tree, Extreme Random Forest, Logistic Regression, Linear Discriminant Analysis (LDA), Quadratic Discriminant Analysis (QDA). In addition, we used a PyTorch \citep{paszke2017automatic} implementation of Shallow Linear NN (Perceptron), and Shallow Max NN.

The results obtained by the best models are depicted in Table \ref{results_image}. In all the cases, we used only 1/4 of the data and achieved best performances. In all cases except one, the best model was able to learn the exact coding scheme and to achieve perfect results. The only exception is the "Multiple Location" scheme, and this is due to the inherent ambiguity in this encoding scheme\footnote{There is a chance of having the exact same value in distant pixels}. We note in passing that in many cases some of the other models achieved comparable results. For example, although random forest is not in the table, in many of the cases, it came very close to the winning model, thus making it a good universal model choice. 

%
%
%

\begin{table}[h]
\small
\begin{center}
\begin{tabular}{|c|c|c|c|}
\hline 
Truth Signal & ConvNet & \multicolumn{2}{c|}{Best Model} \\ 
\hline\hline 
One Pixel & 84.9\% & 100\% & Dec. Tree \\ 
\hline 
Pattern Pixels & 84.9\% & 100\% & Dec. Tree \\ 
\hline 
Random Pixel & 84.9\% & 100\% & Shallow NN \\ 
\hline 
Multiple Locations & 84.9\% & 98.4\% & QDA \\ 
\hline 
Noise & 84.9\% & 100\% & Perceptron \\ 
\hline 
Mean & 84.9\% & 100\% & Shallow NN \\ 
\hline 
\end{tabular} 
\end{center}
\caption{CIFAR10 - Image Classification Accuracy}\label{results_image}
\vspace{-1em}
\end{table}

%
%

The results demonstrate the impact of the ''blindspots''. Each of the truth-revealing signals that
were used corresponds to either local information that is being ``smoothed out'' (one-pixel encoding, pattern pixels), a
global signal that is being discarded (noise encoding) and a moving target (random pixel, multiple locations). Each of these
signals, although sufficient to provide perfect classification, could not be perceived by a convolutional network similar to the one we used.
Each signal may certainly be addressed by changing the network architecture to meet the specific characteristics of that signal 
(e.g., reducing kernel size to capture local information, or enlarging it to capture global signals), but this will require prior knowledge of that signal and will hamper the generic use of ConvNets. We conjecture that this set of signals will introduce the same ''blindspots'' challenges for many of the popular and commonly used ConvNets.

Some signals are naturally captured by a decision tree, leveraging the direct correspondence between the signal (regardless of position and locality) and the target. Other signals are captured by shallow NNs that focus attention more on the global properties of the data, thus revealing information encoded by a global truth-revealing noise.
Most interesting observation occurs in the case of orthogonal ``noise encoding'', where it was demonstrated that while a deep convolutional network did not integrate the global information, a shallow 1-layered fully connected network achieved perfect prediction within a small number of iterations. This is due to the fact that a shallow NN is able to form a global perspective over the input space and learn widely supported (global) properties. Specifically, in the ''noise encoding'' case, the shallow network learned a weight vector corresponding to each of the noise vectors, which suffices to discriminate between the classes.

To further demonstrate this point, we devised images that hold no information except the label. Specifically, we used blank images with ''one pixel'' signal. 
These images were easily classified by the ConvNet since the truth signal had no interference with the image information.
This is not surprising, as ConvNets are known to be able to memorize easily even random labeling of data \citep{zhang2016understanding}, and as such can easily learn this simple rule once give explicitly.
We view this as additional evidence that it is difficult for the network to capture and use low dimensional information when aggregating over a more substantial part of the spatial space that holds information. 

The discardment of local information added to pixel space can also be possibly explained with tools suggested by \cite{shalev2017failures}, as the gradients for weights in convolutional layers will have a low signal-to-noise ratio over a large input image.
 \label{subsec_vision}
 \subsection{Text classification}
 \label{subsec_text}
 To further evaluate and confirm our findings, we devised a set of experiments in the text domain. We used the framework of \cite{zhang2015character}, including the data set and the Torch implementation that they kindly shared. \cite{zhang2015character} designed two 1-dimensional convolutional networks for text classification at the character level.

The two ConvNets share the same architecture and differ in their size (number of neurons). They are both 
Nine layers deep with six convolutional layers and three fully-connected layers.
The alphabet consists of 70 characters (including space). Hence, each input feature has 70 categorical values, which were numerically coded using one-hot-encoding scheme. Overall, there are 1014 different input features.
The ConvNets also include two dropout modules in between the three fully-connected layers to regularize. In our experiments, we used the smaller network implemented by Zhang et al.


 
The convolutional network was trained to classify segments of written text, similarly to their spatial counterpart on images. As both kinds of models (spatial and temporal ConvNets) suffer from the same limitations, we expected to observe the same phenomenon: that truth-revealing signals with certain properties will be ignored by the network, while being extensively used by the other models.

In their experiments, when performance was compared against traditional models, Zhang et al. used logistic regression at the word level: Bag-of-words/ngrams counts and their TFIDF (term-frequency inverse-document-frequency), resp. Since we were interested in analyzing ''blindspots'', we employed the competing models at the character level, thus enabling a fair and ''clean'' competitive analysis.
Zheng et al. conducted their experiments on eight different datasets. In four out of eight (AG News, Sogou News, DBPedia, Yelp Review Polarity), logistic regression came first. In two others (Yahoo! answers, Amazon Review Polarity) logistic regression came second but with a small margin to the ConvNet. In the last two (Yelp Review Full, Amazon Review Full), the ConvNets won by a significant margin. 
We chose to conduct our experiments on the "Yelp Review" and "Amazon Review" datasets as described by \cite{zhang2015character}.

 \dnote{I'm curious: what are the 5 classes?}
 \snote{I don't remember}
We employed three different coding schemes of the ground-truth labels in the text
\begin{itemize}
\item Mnemonic: The first character in each message was changed to reflect the category label, reminiscent of the short mnemonic form of the element name that is used in the machine-readable encoded document. 
\dnote{I didn't understand the second half of sentence.}
\snote{this is a known category encoding scheme. Ig I remember correctly, it is used for example in news messages, e.g., Reuters, etc.}
\item Length encoding: The label was encoded by setting a fixed length for the text in all the messages of the same class. \dnote{how? did you cut the message? wouldn't this hurt performance?}
\snote{I think that I also used padding. As for performances, Look at Table 3 - It didn't really hurt}
\item Pattern char: Encode the label by changing the characters at specific fixed locations to hold the class number (label). In our study, we used three specific character locations (the 50th, 150th, and 200th) for a binary encoding scheme, such that five labels are encoded with the following "bit" assignment. In each selected location either set the character to be 0 or keep it unchanged. 
\end{itemize}

The \emph{Mnemonic} signal is of the same nature as the \emph{One Pixel} encoding in the images. It provides complete information at the character level. 


The \emph{Length encoding} signal is a global signal that is being discarded by the network although it holds the necessary information. This is similar to the \emph{Noise Encoding} signal at
the image examples.

The \emph{Pattern char} signal is a type of low dimensional signal that is being discarded because of the aggregation and smoothing that is done by the network. It is of the same nature as the \emph{Pattern pixels} encoding, however, it provides only partial information.

The results attained with the small ConvNet on a clean data (without signal injection) were in the same ballpark as the ones reported by \cite{zhang2015character}. The network didn't yield any significant improvement in performances when trained on any of the truth-revealing signal injected data sets.

Similarly to the image classification experiments, for traditional modeling, we used scikit-learn implementation. However, unlike the image classification setting, it turned out that for text classification, a Decision Tree is the best model to handle all injected signals at the character level.
The results are depicted in Table \ref{results_text_yahoo}. It also includes the best-cited results of Zhang et al. using ConvNets at the character level and Logistic Regression at the word level.

We can see that the same observations seen in the spatial case are also visible when applying convolutional network on text. We can attribute once again the three types of truth-revealing signals
to either global information being missed (Length encoding) or local information being smoothed and discarded (Mnemonic, Pattern char). 
We can also see the effect of providing partial vs. complete information - Mnemonic signal provide complete information, Length encoding is a little ambiguous (uses space padding to mark the end of the text, but spaces can be found in the body of the text as well). Finally, Pattern char provides partial information (the bit states are encoded with either 0 or the existing char unchanged, which might be 0 as well since 0 represents the space char).

\begin{table}[h]
\small
\begin{center}
\begin{tabular}{|c|c|c|c|}
\hline 
Truth-Signal & Model & Yelp F. & Amazon F. \\ 
\hline\hline 
Mnemonic & DT & 100\% & 100\% \\  
\hline 
Length encoding & DT & 99.80\% & 99.97\%  \\ 
\hline 
Pattern char & DT & 68.66\% & 69.66\%  \\ 
\hline 
Clean data  & LR & 59.86\% & 55.26\% \\ 
\hline 
Clean data & ConvNet & 62.05\%  & 62.05\%  \\ 
\hline 
\end{tabular}
\end{center}
\caption{Text Classification accuracy}
\label{results_text_yahoo}
\vspace{-2em}
\end{table}

\section{Discussion}
\label{sec_discussion}
In this paper, we introduced the notion of ''blindspot'' as a mean to study the strength and limitations of convolutional networks. 
Our method is based on injecting a truth-revealing signal to the data, indiscernible to the network, 
thus hitting time and again the network's blind spots. The signal does not impair the network's existing performances. Instead, it provides an opportunity for a significant performance boost by models that can capture it.

The models we used to compare with are all standard, integral part of commonly used generic Machine Learning toolboxes.
In all our experiments we used the Python's Scikit-learn package \citep{scikit-learn} in a straightforward manner, 
namely no special data pre-processing, no code enhancements, nor any tailor-made augmented functionality.

We conducted our series of experiments in two different application domains, 
images and text, using two different convolutional networks that differ in their input space encoding (discrete symbols for text, continuous pixel values for image)
and their processing over data (temporal vs. spatial convolution).
Still, we ended up with similar observations and similar ''blindspots''.
This demonstrates the robustness of our findings, and it enables us to come up with a unified set of insights and recommendations.

In designing the studies, we started with analyzing the principle characteristics of ConvNets, hypothesize on their deficiencies,
and then designed experiments to either validate or disprove these assumptions. 
Our leading theme is that ConvNets are less susceptible to complete and incomplete information with the following characterization 
\begin{itemize}
\item 
Dimension - The information resides either at a very low dimension or spreads over very many dimensions 
\item
Global - When the information is in the form of a global property, such as max or some forms of summary statistics
\end{itemize}

Taking advantage of these deficiencies, we were able to layout simple design principles for the truth-revealing signals, 
which are manifested by the following groupings of the encoding schemes
\begin{enumerate}
\item 
Low dimensional signals - One-pixel encoding, Pattern pixel encoding, Pattern char encoding, Mnemonic encoding
\item High dimensional signal - Noise encoding, Length encoding
\item Stochastic signal - Random pixel encoding, Multiple locations
\end{enumerate}

The various signals injected to the data are artificially made and highly unlikely to appear in natural data. 
Still, they can be seen as extreme cases of naturally occurring situations. For example, the \emph{orthogonal noise} we've added to the image dataset
that was missed by the ConvNet, can be seen as an extreme case of global noise that is informative. A natural case of this kind of noise is when a global signal like
physical aberrations from the photography machinery is added to images and is correlated with the conditions in which an image was taken. It is thus informative to the classification task,
and ignoring it can hurt potential performance. 
A similar example can be seen in text, where the \emph{length encoding} we have shown to be ignored, can represent a global structural attribute of the text lines. For example, the style that at which a line was written.

An interesting observation is that properties, such as invariance to shift and scale, which are usually considered as an advantage of ConvNets, may also be harmful, e.g., when the information is coded in the shift and scale itself. 

In some cases, a shallow network was able to capture the (important) information that a deep network failed to notice. 
In another case, when we removed all the information except the truth-revealing signal, the deep network was able to capture it and reached perfect performance. We believe that these two phenomena are rooted at the same characteristic of the deep network: the form of the interplay that deep ConvNets conduct between feature construction (convolution layers) and feature selection (pooling layers).
By shutting down obscuring information, and by using a wider fan-in (using shallow networks), we manage to intervene this interplay and overcome the difficulties in some of the cases. 

The comparison of a shallow vs. deep network is of particular interest since expressiveness arguments suggest that deep neural networks provide a universal advantage over shallow networks \citep{cohen2015expressive}. 
Our findings are not necessarily constructing, as to modify the models used, but it seems that there is a need for a finer characterization of the strength of deep architectures, and how it is manifested in various domains.

Using ConvNets on a new source of data should take into account the ''blindspots'' presented in this work.
This is especially important when the data is pre-processed to suit convolutional processing, possibly incurring noise artifacts or ignoring crucial information. 
In our studies we witness this phenomenon with the text example, where the one-hot-encoding fed into the networks allows us to insert the \emph{length encoding} signal without being noticed. The same phenomena might occur for example, when learning
 from audio signals by running a ConvNets on spectrograms \citep{schluter2014improved} and ignoring valuable information.


Last but not least is the utilization of traditional machine learning models - Compared with ConvNets, these models offer a significant computational advantage, require far fewer data to converge, do not involve a lot of parameter tuning, and yield simpler hypotheses, which 
should be favored when performance is comparable.

\bibliography{cnn_blindspots}
\bibliographystyle{nips-no-url}

\end{document}